\pdfoutput=1

\documentclass[11pt]{article}

\usepackage[]{acl}

\usepackage{times}
\usepackage{latexsym}
\usepackage{subfigure}
\usepackage{tabularx}
\usepackage{booktabs}
\usepackage{enumitem}%
\usepackage[T1]{fontenc}

\usepackage[utf8]{inputenc}

\usepackage{microtype}
\usepackage{inconsolata}

\usepackage{multirow}
\usepackage{balance}
\usepackage{verbatim}
\usepackage{diagbox}
\usepackage{xcolor}
\usepackage{bm}
\usepackage{amsmath}
\usepackage{graphicx}
\usepackage[most]{tcolorbox}
\usepackage[normalem]{ulem}
\useunder{\uline}{\ul}{}

\title{Depression Detection on Social Media with Large Language Models}

\author{
  Xiaochong Lan$^{1}$\thanks{Equal contribution.} \ \ \ 
  Zhiguang Han$^{2}$\footnotemark[1] \ \ \ 
  Yiming Cheng$^{3}$ \ \ \ 
  Li Sheng$^1$ \\
  \textbf{Jie Feng}$^1$ \ \ \ 
  \textbf{Chen Gao}$^{1}$\thanks{Corresponding author.} \ \ \ 
  \textbf{Yong Li}$^{1}$\footnotemark[2] \\
  $^1$Department of Electronic Engineering, BNRist, Tsinghua University \\
  $^2$College of Computing and Data Science, Nanyang Technological University \\
  $^3$Department of Computer Science, The University of Chicago \\
  \small \texttt{lanxc22@mails.tsinghua.edu.cn} \ \ \ \ 
  \texttt{\{chgao96, liyong07\}@tsinghua.edu.cn}
}

\begin{document}
\maketitle
\begin{abstract}
Limited access to mental healthcare resources hinders timely depression diagnosis, leading to detrimental outcomes.
Social media platforms present a valuable data source for early detection, yet this task faces two significant challenges: 1) the need for medical knowledge to distinguish clinical depression from transient mood changes, and 2) the dual requirement for high accuracy and model explainability.
To address this, we propose DORIS, a framework that leverages Large Language Models (LLMs).
To integrate medical knowledge, DORIS utilizes LLMs to annotate user texts against established medical diagnostic criteria and to summarize historical posts into temporal \textit{mood courses.}
These medically-informed features are then used to train an accurate Gradient Boosting Tree (GBT) classifier.
Explainability is achieved by generating justifications for predictions based on the LLM-derived symptom annotations and mood course analyses.
Extensive experimental results validate the effectiveness as well as interpretability of our method, highlighting its potential as a supportive clinical tool.
\end{abstract}

\section{Introduction}
Depression is a major global health challenge, yet significant barriers like cost, stigma, and lack of infrastructure prevent many from receiving timely care, creating a critical treatment gap~\cite{worldint, khan2016economic, rusch2005mental}. To help bridge this gap, analyzing public social media data offers a scalable approach for early detection. By examining patterns in users' posts, data-driven systems can provide clinicians with timely signals of potential distress, augmenting traditional diagnostic methods~\cite{orabi2018deep,shen2017depression,sarkar2022predicting}.

However, building a clinically valid and trustworthy system from this data presents significant hurdles. Many existing deep learning methods~\citep{lin2020sensemood,orabi2018deep,sarkar2022predicting} lack a systematic integration of medical knowledge, struggling to distinguish the complex patterns of clinical depression from transient sadness~\cite{zhang2023phq}. Furthermore, they often function as ``black boxes'', failing to provide the transparent, evidence-based reasoning that clinicians require for validation and trust. While Large Language Models (LLMs) offer powerful interpretation capabilities~\cite{yang2023towards}, their direct use for classification can lead to lower accuracy and reliability issues~\cite{arcan2024assessment,hua2024large}. This reveals a critical need for a framework that synergistically combines clinical knowledge, robust prediction, and explainability.

\begin{figure}[t]
\centering
\includegraphics[width=7.6cm]{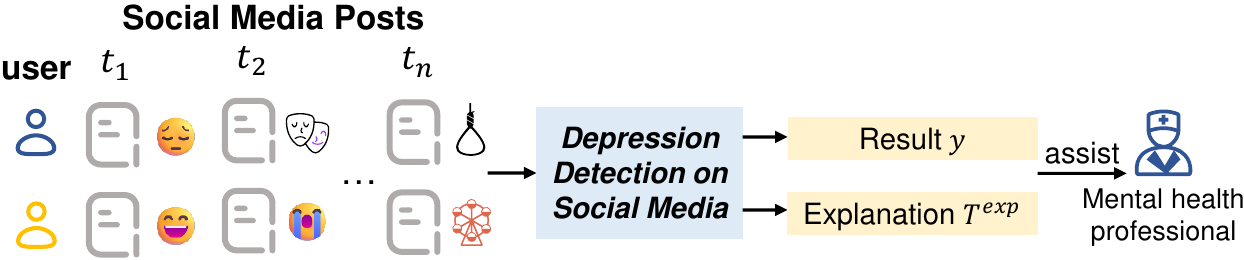}
\caption{Depression detection on social media analyzes users' social media post history to identify potential depression, offering insights to support mental health professionals in depression assessment.}
\label{fig:problem}
\vspace{-0.5cm}
\end{figure}

To address these challenges, we propose DORIS (short for \textbf{D}ia-gn\textbf{O}stic C\textbf{R}iteria-Guided Mood H\textbf{IS}tory-Aware). For the first challenge, we leverage LLMs' strong language understanding capabilities to recognize complex expressions of depression symptoms based on DSM-5 criteria. We also systematically model users' mood courses by filtering emotionally intensive posts and using LLMs to analyze their temporal patterns, enabling holistic assessment of users' emotional fluctuations. For the second challenge, we design a hybrid approach that combines LLMs for explainable symptom and mood course analysis with GBT classifier for a robust final prediction. This design achieves high accuracy through the classifier while maintaining explainability in two ways: through concrete evidence derived from medical knowledge (annotated symptoms and mood courses), and through LLM-generated explanations that connect this evidence to depression assessment. 

Our method shows significant improvements in experimental evaluation, achieving a 0.0361 gain in AUPRC over the current best baseline. Ablation studies validate each component's contribution, and case studies demonstrate the system's ability to generate explainable assessments backed by concrete evidence.

Our contributions can be summarized as follows:
\begin{itemize}[leftmargin=*,itemsep=0pt,parsep=0.2em,topsep=0.0em,partopsep=0.0em]
    \item We propose a novel hybrid depression detection framework that synergistically combines a robust classifier for accurate prediction with an LLM for explainable, clinically-grounded feature generation, addressing the critical dual need for accuracy and interpretability.
    \item We introduce a methodology to systematically operationalize medical knowledge by using LLMs for both fine-grained DSM-5 symptom annotation and longitudinal mood course analysis, creating features that align closely with clinical diagnostic practices.
\item We demonstrate through extensive experiments on real-world data that our approach not only significantly improves detection performance but also provides clinically interpretable explanations grounded in medical knowledge, demonstrating a viable pathway for creating practical AI-driven tools that support clinicians and improve mental healthcare access in low-resource settings.
\end{itemize}

\begin{figure*}[t]
\centering
\includegraphics[width=15cm]{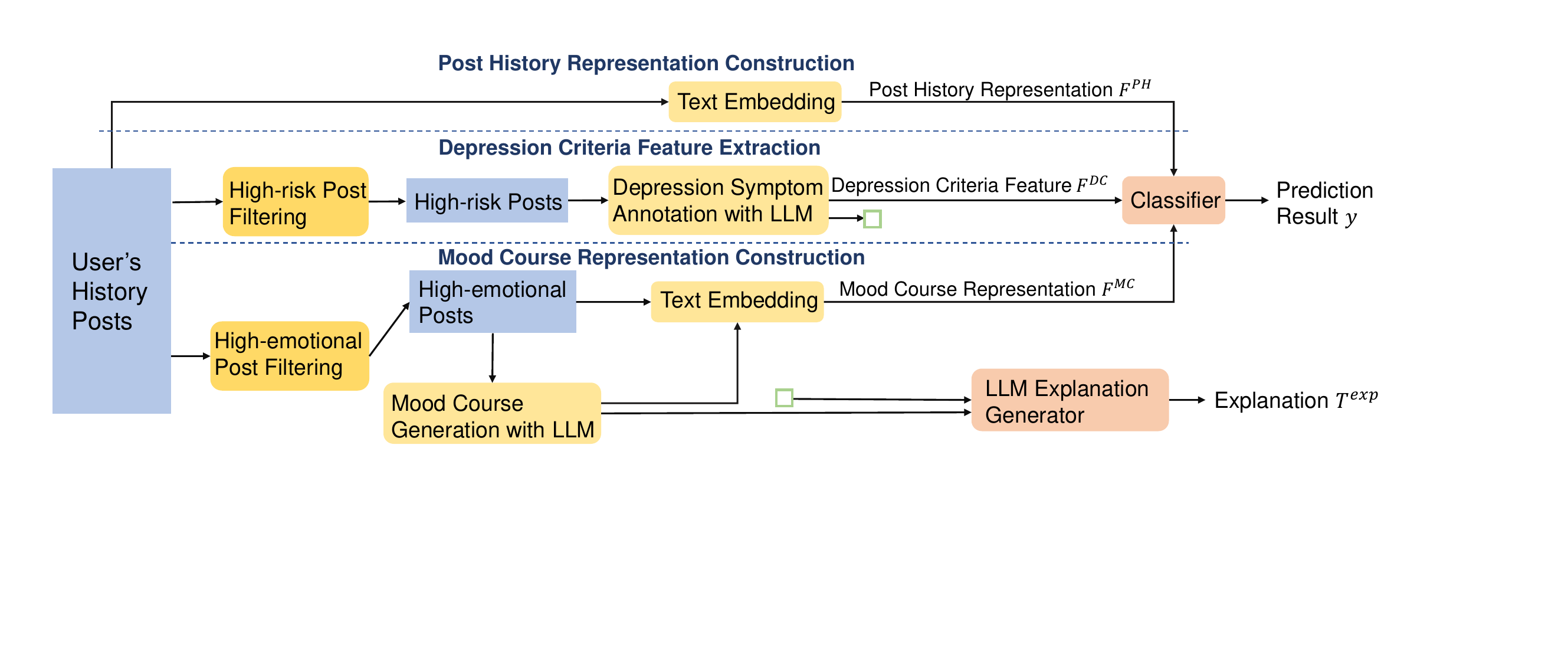}
\caption{Illustration of DORIS. Through the collaboration of the LLM and the text embedding model, we obtain three key features: depression symptom feature, post history representation, and mood course representation. The classifier uses these three features to make its judgment; the LLM uses annotations of depression symptoms and descriptions of the mood course to generate explanation for the system's decision.} 
\label{fig:model}
\end{figure*}

\section{Methodology}
\subsection{Problem Formulation}
This study addresses the task of depression detection based on users' social media posts. Users often use posts to document events around them and express their feelings, making these candid expressions valuable for gauging the likelihood of depression. Given a user $u$, let $P=\{p_1, p_2, ..., p_n\}$ denote their historical posts with the corresponding timestamps $\{t_1, t_2, ..., t_n\}$. Our objective is to learn a mapping function $f: P \rightarrow y$ that predicts the depression status $y \in \{0,1\}$ of user $u$, where $y=1$ indicates the presence of depression and $y=0$ indicates its absence.

In the following part, we describe our proposed DORIS in detail. The architecture of DORIS is illustrated in Figure~\ref{fig:model}.

\subsection{Diagnostic Criteria Feature Extraction}

\subsubsection{Annotation with LLM} Clinical practice offers well-established diagnostic frameworks for depression that have proven highly effective in professional assessment. To bring this medical expertise to social media-based detection, we incorporate standardized diagnostic criteria into our system's design. Specifically, we utilize DSM-5~\citep{regier2013dsm}, a widely recognized diagnostic tool in psychiatry that provides comprehensive criteria for mental disorder assessment and diagnosis. The core criteria in DSM-5, which guide mental health professionals in their diagnostic practice, are detailed in Table~\ref{tab:dsm5}. 

\begin{table}[h!]
\centering
\caption{A concise summary of the symptoms of depression as defined in the DSM-5. To be diagnosed with depression, five (or more) of the listed symptoms must be present during the same two-week period.}
\begin{tabular}{|l|}
\hline
A. Depressed mood \\
B. Loss of interest/pleasure \\
C. Weight loss or gain \\
D. Insomnia or hypersomnia \\
E. Psychomotor agitation or retardation \\
F. Fatigue \\
G. Inappropriate guilt \\
H. Decreased concentration \\
I. Thoughts of suicide \\ \hline
\end{tabular}
\label{tab:dsm5}
\end{table}

Prior attempts to integrate clinical standards via DSM-5 criteria or its simplified screening tool PHQ-9~\cite{yadav2020identifying,nguyen2022improving,zhang2023phq} rely on simple classification models that struggle to comprehend complex symptom expressions in social media texts. The emergence of large language models (LLMs) offers a promising solution to this challenge. LLMs exhibit a remarkable capacity for semantic understanding~\citep{achiam2023gpt}, and research has demonstrated their potential to replace human annotators in certain tasks~\citep{ziems2023can}. Here, we employ LLMs to accurately annotate texts, specifically to identify if and which self-expressed symptoms of depression are present in posts. The prompt is described in Appendix~\ref{sec:prompt}.

By post-processing the output from the LLM, we can generate a 9-dimensional vector for each post, with each element being 0 or 1, indicating the absence or presence of a specific depression symptom. For instance, if the output for a post $p$ is \textit{(G, I)}, then the corresponding vector $E_p$ would be $(0, 0, 0, 0, 0, 0, 1, 0, 1)$.

\subsubsection{Efficient Annotation.}
\label{sec:efficient_annotation}
The use of LLMs incurs substantial financial and energy costs. Since the vast majority of posts on social media platforms are unrelated to depression, annotating all posts would be exceedingly wasteful. To address this issue, we design an efficient annotation approach that first filters for high-risk texts through carefully validated symptom templates (detailed in Appendix~\ref{sec:symptom_templates}) and then annotates only those texts with LLMs.
We developed symptom templates through a four-stage process. First, psychiatrists created initial templates from DSM-5 diagnostic criteria. Second, we refined them using a corpus of 500 social media posts independently annotated by four clinicians, evaluating inter-annotator agreement (Cohen's kappa) to improve them. Third, experts iteratively reviewed the templates for clinical accuracy and coverage until reaching consensus. Finally, experimental validation confirmed that filtering with our templates achieves comparable performance to full LLM annotation at a significantly lower computational cost.

Then, we get the text embedding of all symptom templates using a text embedding model:
\begin{equation}
    H_i=\text{Encoder}(T^{D C}_i), \text{for}\ i =\text{A}\ \text{to}\ \text{I}. 
\end{equation}
For each post $p$, we compute its embedding as :
\begin{equation}
H_p =\text{Encoder}(p).
\end{equation}
Next, we calculate the average similarity between post $p$ and each symptom template:
\begin{equation}
Sim_p=\text{mean}(Sim(H_p,H_i)), \text{for}\ i =\text{A}\ \text{to}\ \text{I},
\end{equation}
where $Sim_p$ represents the depression risk level of post $p$. We only use the LLM to annotate posts with the top $k\%$ of $Sim_p$ scores, while the depression symptom vector $E_p$ for all other posts is directly set to a zero vector. Here $k$ is a hyperparameter. Finally, for each user $u$, we average all their depression symptom vectors to obtain their diagnostic criteria feature $F_u^{D C}$:
\begin{equation}
F_u^{D C}=\frac{1}{N} \sum_{p=1}^N E_p,
\end{equation}
where $N$ is the total number of posts by user $u$.

\subsection{Mood Course Representation Construction}
The mood course, also termed mood trajectory in clinical literature, represents the temporal pattern and progression of emotional states~\citep{cochran2016data} and is critical in diagnosing clinical depression~\citep{costello2002development}. It delineates the onset, duration, and recurrence of mood episodes, providing insights into the disorder's nature and trajectory~\citep{frias2017longitudinal}. Accurately modeling mood course is pivotal for distinguishing depressive disorders from transient mood fluctuations, facilitating early detection and appropriate intervention~\citep{rudolph2006mood}. However, former works on depression detection have largely overlooked the mood course, focusing instead on static mood snapshots. Our study bridges this gap by explicitly modeling mood course and integrating it into our classification system. Next, we detail our approach.

\subsubsection{Posts Filtering} Not all posts are emotionally charged. We begin by filtering posts with a high emotional content. Following \citet{oatley1987towards}, we categorize emotions into five main types: 1) anger, 2) disgust, 3) anxiety, 4) happiness, and 5) sadness. For each of these emotional categories, with the help of professional psychologists, we establish a template of emotional expressions (detailed in Appendix~\ref{sec:emotion_templates}) using the same systematic four-stage process described for depression symptom templates.
For instance, the template for sadness, $T_5^E$, is defined as:

\begin{tcolorbox}[
    width=0.95\linewidth,
    colback=black!5!white,  %
    boxrule=0pt
    ]
\textit{"I am sad, sorrowful, melancholic, in pain, lost, depressed, pessimistic, tearful, grieving, mournful, depressed, suicidal, heartbroken, devastated, upset, crying, deeply saddened, disconsolate, dejected, lamenting, desolate, gloomy, mournful, weeping bitterly, desperate, heartbroken, indignant."}
\end{tcolorbox}

For each emotion template, we generate a representation using a pre-trained embedding model:
\begin{equation}
H^E_j=\text{Encoder}(T_i^E), \text{for}\ j\ = 1,2,...,5.
\end{equation}

\subsubsection{Representation Construction}
For each post $p$, we obtain its embedding $H_p$. We calculate the similarity between post $p$ and each emotion template as:
\begin{equation}
Sim_{p j}=Sim(H_p,H_j), \text{for}\ j\ = 1,2,...,5.
\end{equation}
For each emotion $j$, we retain posts within the top $m\%$ of similarity, forming the set $S_j$. Here $m$ is a hyperparameter. The final set of posts with high emotional content, $S$, is the union of all $S_j$.

For each user $u$, we intersect their historical post set $P_u$ with the high emotional content set 
$S$ to obtain $P_u^E$. Based on this subset of emotionally expressive posts, we use an LLM to synthesize a description of user $u$'s mood course, $T^{M C}$, the prompt is described in Appendix~\ref{sec:prompt}.

We then compute the embedding of $T^{M C}$:
\begin{equation}
H^{M C}=\text{Encoder}(T^{M C}).
\end{equation}
The user $u$'s mood course representation $F_u^{M C}$ is computed as:
\begin{equation}
F_u^{M C}=\alpha H^{M C} + \beta  \frac{1}{\left|P_u^E\right|}\sum_{p \in P_u^E} H_p,
\end{equation}
where $\left|P_u^E\right|$ is the total number of posts in $P_u^E$, and $\alpha$, $\beta$ are hyperparameters. Here we get $F_u^{M C}$ as a comprehensive representation of the types and evolution of user historical moods.

\subsection{Post History Representation Construction}

In the sections above, we construct the diagnostic criteria feature and mood course representation from users' historical posts through filtering and labeling. While these filtering steps emphasize aspects crucial for medical diagnosis of depression, they may also result in information loss. To address this, we construct a representation of the user's post history as follows:
\begin{equation}
F^{P H}=\frac{1}{N}\sum_{p=1}^N H_p,
\end{equation}
where $H_p$ is the embedding of the $p$-th post, and $N$ is the total number of posts by the user.
\subsection{Training and Predicting}
In this section, we describe our training and predicting methodology. First, we integrate the various features. $F^{M C}$ and $F^{P H}$, which shares the same space, are directly summed to avoid increasing the dimension and exacerbating the risk of overfitting. Conversely, $F^{D C}$ resides in a distinct space, and thus we concatenate this feature with the sum of the first two parts. The final feature vector $F$ is obtained as:
\begin{equation}
F=\operatorname{Concat}\left(F^{M C}+F^{P H}, F^{D C}\right)
\end{equation}
We employ Gradient Boosting Trees (GBT) for classification. GBT is an ensemble learning method that constructs a sequence of decision trees, where each subsequent tree aims to correct the errors of its predecessor. It excels in automatically performing feature interactions by selecting optimal split criteria within its decision trees, thus effectively fusing the components of 
$F$. The final prediction, denoted as $y$, is a binary classification result derived from the ensemble model. The process is formalized as follows:

First, the ensemble model $G$ is initialized and then enhanced iteratively by adding decision trees:
\begin{equation}
G_m(x)=G_{m-1}(x)+\nu \cdot h_m(x).
\end{equation}
Each tree $h_m(x)$ is fitted to the negative gradient of the loss function evaluated at $G_{m-1}$, aiming to minimize:
\begin{equation}
\sum_{i=1}^N L\left(y_i, G_{m-1}\left(x_i\right)+h_m\left(x_i\right)\right).
\end{equation}
The prediction y is given by the sign of $G_M(x)$, the output after $M$ iterations:
\begin{equation}
y=\operatorname{sign}\left(G_M(x)\right).
\end{equation}
Here, $L$ represents the loss function, $N$ is the number of samples, $M$ is the total number of trees, and $\nu$ is the learning rate.

Furthermore, for safety-critical tasks with severe class imbalance like depression detection, ensuring the model's output probabilities are reliable is crucial. The raw scores from a GBT model may be miscalibrated. To address this, a post-processing calibration step can be applied to improve clinical utility. Methods like Isotonic Regression can adjust the classifier's scores to better reflect true posterior probabilities. This helps correct for potential biases, such as underestimating the probability of the minority class, thereby providing clinicians with more trustworthy confidence scores.

\subsection{Design for Explainability}
Explainability is critical for safety-sensitive tasks like depression detection~\citep{zhang2022natural}. While Large Language Models (LLMs) show potential for generating interpretable analysis~\citep{yang2023towards,yang2023mentalllama}, their direct use for classification can be unreliable and sensitive to prompting~\citep{hua2024large}. Our hybrid approach is explicitly designed to overcome this by combining a robust traditional classifier with the interpretive power of an LLM, leveraging the strengths of both.

Our system's explainability is two-fold. First, the final prediction is based on two clinically-relevant, human-readable features generated by an LLM: 1) a set of annotated symptoms aligned with DSM-5 criteria, and 2) a narrative summary of the user's longitudinal ``mood course''. These features serve as direct evidence for the model's reasoning. Second, to synthesize this evidence for the end-user, the system uses an LLM to generate a final explanatory text, $T^{Exp}$, that explicitly connects the identified symptoms and mood patterns to the classification result ($y$). The prompt used for this generation is detailed in Appendix~\ref{sec:prompt}.

\section{Experimental Setup}
\subsection{Implementation Details}
\label{sec:implementation}
Following~\citet{orabi2018deep}, we utilize cosine similarity to calculate the similarity between embeddings. To enhance the usability of our method in low-compute resource settings, we employ a low-resource-demanding pre-trained text embedding model, gte-small~\cite{li2023towards}, which operates smoothly with just 1GB of memory. The text embedding model in our system can easily be switched to other higher-performance models to further improve performance. For the LLM, we use GPT-4o mini~\footnote{https://platform.openai.com/docs/models/GPT-4o-mini}, which requires only an internet connection to interact with it through the API service provided by OpenAI. We have verified that OpenAI's terms of service prohibit storing or retaining any user-provided text data, ensuring data privacy. The LLM in our system can also be substituted with open-source models deployed locally, such as Mentallama~\citep{yang2023mentalllama}, to further ensure data privacy. Overall, our system can run at low computational costs, enhancing its accessibility. We utilize XGBoost~\citep{chen2016xgboost} for an efficient implementation of Gradient Boosting Trees. We run each experiment 5 times and report the average performance across these runs.

\begin{table*}[]
\centering
\small
\caption{Performance of DORIS and baselines on the SWDD dataset. The best scores are in bold, and second best scores are underlined.}
\vspace{-0.3cm}
\begin{tabular}{ccccccc}
\hline
\textbf{Category} & \textbf{Method} & \textbf{Precision} & \textbf{Recall} & \textbf{F1-score} & \textbf{AUROC} & \textbf{AUPRC} \\ \hline
Traditional Method               & TF-IDF+XGBoost & 0.3644          & 0.4312          & 0.3945          & 0.9023          & 0.4303          \\ \hline
\multirow{3}{*}{Deep Learning-Based Methods} & HAN            & 0.5702          & 0.6524          & 0.6075          & 0.8929          & 0.5864          \\
                             & Mood2Content   & 0.7216          & 0.6996         & {0.7106}    & {0.9537}    & {0.7774}    \\ 
                             & AMM-Net       & 0.6851          & 0.6861          & 0.6856          & 0.9220          & {\ul 0.7786}          \\ \hline
\multirow{4}{*}{PLM-Based Methods}    
& FastText       & {\ul 0.7467}    & 0.5586          & 0.6421          & 0.9441          & 0.6255          \\
&gte-small &0.7359 &0.6524 &0.6916 &0.9499 &0.6959
\\
                             & BERT           & 0.6667          & 0.6294          & 0.6531          & 0.9481          & 0.7102          \\
                             & MentalRoBERTa  & 0.7326          & 0.6272          & 0.6774          & 0.9423          & 0.6880          \\ \hline
\multirow{2}{*}{Medical Knowledge-Guided Methods} & PHQ9 (Score)   & 0.7137        & 0.6974        & {\ul 0.7055}        & {\ul 0.9522}        & 0.7703        \\
                             & PHQ9 (Vector) & 0.7221          & 0.6852          & 0.7032          & 0.9479          & 0.7631          \\ \hline
\multirow{2}{*}{LLM-Based Methods}      & GPT-4o mini      & 0.0895          & 0.7122          & 0.1590          & 0.6603          & 0.0767          \\
                             & MentalLLama    & 0.0899          & {\ul 0.7780}    & 0.1612          & 0.6821          & 0.0811          \\ \hline
Our Method                                    & DORIS          & \textbf{0.7606} & \textbf{0.7902} & \textbf{0.7750} & \textbf{0.9722} & \textbf{0.8147} \\ \hline
\end{tabular}
\vspace{-0.2cm}
\label{tab:main}
\end{table*}

\subsection{Dataset}
Our primary study utilizes a cohort from the ethically-sourced SWDD dataset~\cite{cai2023depression}, consisting of \textbf{1,000} expert-identified depressed users and \textbf{19,000} control users to reflect a real-world 1:19 prevalence ratio. Following best practices~\cite{eichstaedt2018facebook} and accommodating model constraints, we retained only the final six months of each user's post history. This resulted in a corpus of approximately 70k posts from the depressed group and 1.3M from the control group. To further validate the generalizability of our approach, we also conducted experiments on the Twitter Mental Disorder dataset~\footnote{https://www.kaggle.com/datasets/rrmartin/twitter-mental-disorder-tweets-and-musics.}. For a more detailed description of the datasets, please refer to Appendix~\ref{sec:dataset}.

\subsection{Baseline Methods}
We employ various baselines, including methods combining traditional feature extraction with classifiers: TF-IDF+XGBoost~\citep{ramos2003using,chen2016xgboost,wu2023exploring}, deep learning approaches: HAN~\citep{yang2016hierarchical}, Mood2Content~\citep{cai2023depression}, and AMM-Net~\citep{sarkar2022predicting}, PLM-based methods: FastText~\citep{joulin2016fasttext}, BERT~\citep{devlin2018bert}, gte-small~\cite{li2023towards} and MentalRoBERTa~\citep{ji2023domain}, medical knowledge-guided methods: PHQ9 (Score) and PHQ9 (Vector)~\cite{nguyen2022improving}, as well as LLM-based methods like MentalLLama~\citep{yang2023mentalllama} and GPT-4o mini. A more detailed description of comparison methods is in Appendix~\ref{sec:baseline}.

\subsection{Evaluation Metrics}

Following prior works, we evaluate our method using five metrics: Precision, Recall, F1, AUROC, and AUPRC. AUPRC can be considered as the most important metric, as it best reflects the classifier's performance on highly imbalanced datasets~\citep{davis2006relationship}.

\section{Experimental Analysis}
We conducted extensive experiments to address the following research questions:
\begin{itemize}[leftmargin=*]
    \item \textbf{RQ1:} How does our proposed system, DORIS, perform on the depression detection task compared to state-of-the-art methods?
    \item \textbf{RQ2:} How effective is DORIS at generating clinically relevant and evidence-based explanations for its predictions?
    \item \textbf{RQ3:} To what extent does each component of DORIS contribute to its overall performance?
    \item \textbf{RQ4:} How do key hyperparameter settings impact the performance of our method?
\end{itemize}

Due to space constraints, we present the experimental results for RQ4 in the Appendix.

\subsection{Overall Performance (RQ1)}
The results in Table~\ref{tab:main} demonstrate the superiority of our approach on the SWDD dataset. DORIS outperforms all baselines across all metrics, achieving an absolute improvement of 0.0361 in AUPRC over the strongest baseline, AMM-Net. This performance gain stems from our hybrid design, which addresses the weaknesses of alternative methods.

Our experiments confirm that directly using LLMs like GPT-4o mini for classification leads to poor performance, exhibiting high recall but very low precision. Furthermore, while other methods incorporate medical knowledge (e.g., PHQ9-based models), their reliance on less capable backbones like BERT limits their ability to understand complex symptom descriptions. In contrast, DORIS leverages a powerful LLM to effectively model clinical criteria and mood courses. By using the LLM for sophisticated, clinically-grounded feature engineering rather than direct prediction, our task-specific design effectively combines the strengths of modern language models and robust classifiers, leading to state-of-the-art results.

We also conduct experiments on the Twitter Mental Disorder dataset to further validate the effectiveness and generalizability of DORIS. The results are presented in Table~\ref{tab:additional_results}.

\begin{table}[h!]
\centering
\caption{Performance comparison on the Twitter Mental Disorder Dataset.}
\label{tab:additional_results}
\resizebox{\columnwidth}{!}{%
\begin{tabular}{cccc}
\hline
\textbf{Model} & \textbf{F1-score} & \textbf{AUROC} & \textbf{AUPRC} \\
\hline
TF-IDF+XGBoost & 0.9156 & 0.7149 & 0.9675 \\
FastText & 0.9189 & 0.7879 & 0.9770 \\
Mood2Content & 0.9101 & 0.8112 & 0.9775 \\
gte-small & 0.9028 & 0.8089 & 0.9812 \\
BERT & 0.9211 & 0.7581 & 0.9753 \\
MentalRoBERTa & 0.9211 & 0.8143 & 0.9824 \\
\hline
DORIS & \textbf{0.9296} & \textbf{0.8647} & \textbf{0.9878} \\
\hline
\end{tabular}%
}
\end{table}

As the results show, DORIS consistently outperforms strong baselines on this additional dataset across all metrics. This reinforces the findings from the SWDD dataset and demonstrates the robustness of our approach across different data sources.

\subsection{DORIS's explainability (RQ2)}
\begin{figure}[t!]
\centering
\includegraphics[width=\columnwidth]{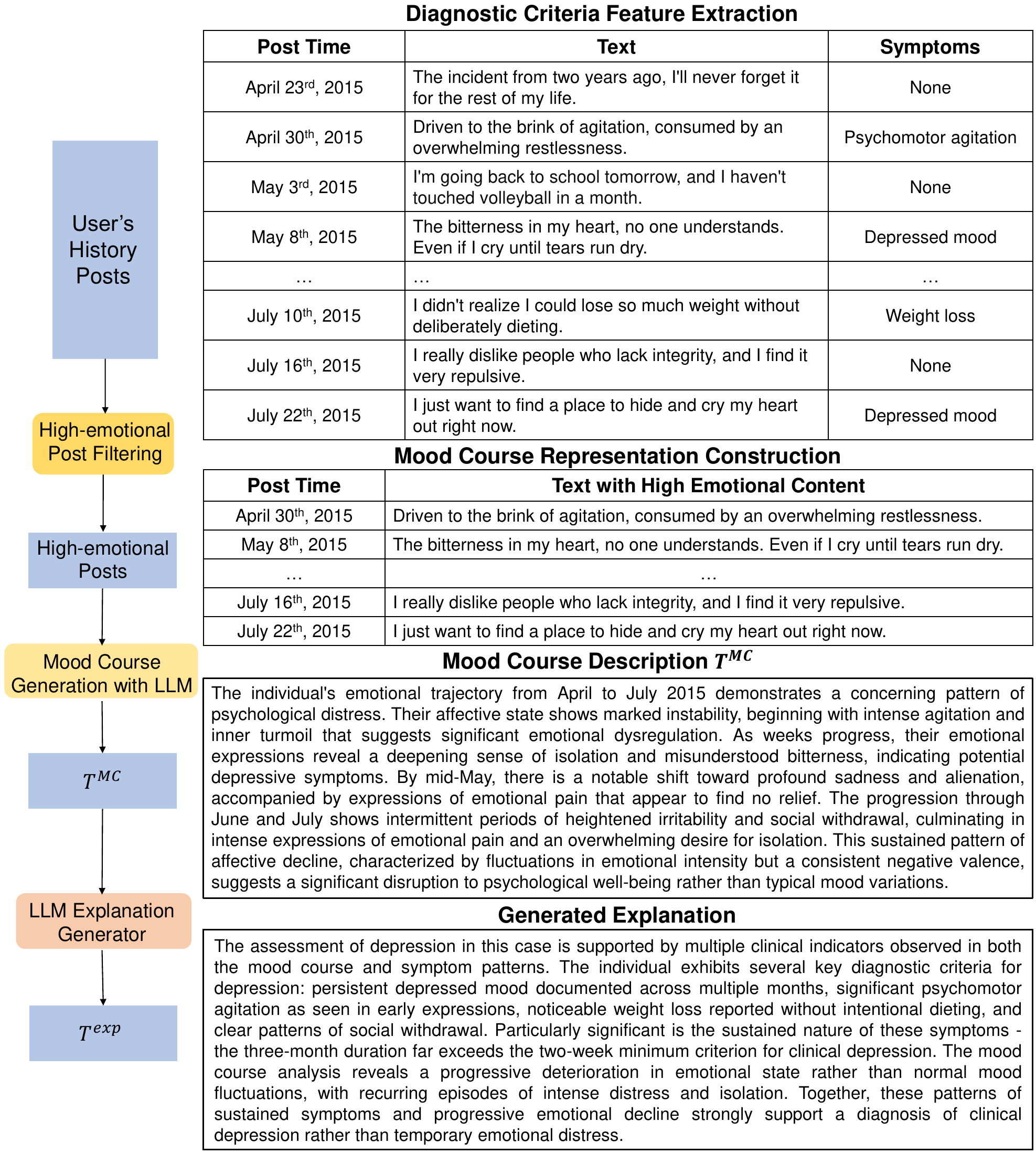}
\caption{A case study of DORIS's output.}
\label{fig:case}
\end{figure}

In our system, we annotate symptoms exhibited in users' posts, summarize the users' mood course, and based on these, provide explanations for the system's final decision output. An example is presented in Figure~\ref{fig:case} to aid in better understanding the operation of our system. As seen, during the diagnostic criteria feature construction stage, our method accurately identifies depression-related symptoms in posts. In the mood course representation construction stage, our approach selects posts with high emotional density, thereby generating a mood course description. By combining identified symptoms of depression and the user's mood course, the system can generate evidence-supported explanations for its judgment. The DORIS system's accurate annotations of depression symptoms, detailed analysis of mood course, and the final evidence-supported explanations are all visible to mental health professionals. This allows the system to serve as a supportive clinical tool in depression assessment.

\subsection{Ablation Study (RQ3)}

\begin{table}[t]
\centering
\caption{Experimental results of ablation study.}
\label{tab:ablation}
\vspace{-0.2cm}
\resizebox{\columnwidth}{!}{%
\begin{tabular}{lccc}
\hline
& \textbf{F1-score} & \textbf{AUROC} & \textbf{AUPRC} \\ \hline
Full Design & \textbf{0.7750} & \textbf{0.9722} & \textbf{0.8147} \\
w/o Diagnostic Criteria Feature & 0.6867 & 0.9679 & 0.7739 \\
w/o LLM Annotation & 0.7028 & 0.9685 & 0.7861 \\
w/o Mood Course Representation & 0.7415 & 0.9660 & 0.7932 \\
w/o Mood Course Summary & 0.7635 & 0.9699 & 0.8035 \\
w/o Post History Representation & 0.7345 & 0.9660 & 0.7817 \\ \hline
\end{tabular}%
}
\end{table}

In our method, the GBT classifier utilizes three constructed features: diagnostic criteria feature, mood course representation, and post history representation. We conduct ablation studies by removing each feature and observing the performance changes. The results are presented in Table~\ref{tab:ablation}.

The results clearly indicate that the diagnostic criteria feature is the most critical component. Its removal causes the most substantial performance drop, with AUPRC decreasing from 0.8147 to 0.7739. This underscores the crucial role of integrating established medical knowledge. Within this module, the LLM-based symptom annotation is also vital; replacing it with a simpler template-matching approach (w/o LLM Annotation) still results in a significant AUPRC drop to 0.7861, demonstrating the LLM's superior ability to capture nuanced symptom expressions.

The post history representation, which provides a holistic view of a user's activity, is also highly impactful. Removing it leads to the second-largest performance decline (AUPRC of 0.7817), confirming its importance in capturing contextual information that more targeted features might miss. Finally, the mood course representation proves essential for modeling temporal emotional patterns. Its absence reduces the AUPRC to 0.7932, with its sub-component, the LLM-generated summary, also showing a clear contribution.

These results collectively demonstrate that each component of DORIS makes meaningful contributions to its overall performance, confirming the effectiveness of every part of DORIS's design.

\section{Conclusion and Future Work}
\label{sec:conclusion}
In this work, we present \textit{DORIS}, a novel depression detection system that achieves high accuracy and interpretability by aligning with clinical diagnostic practices. The system integrates the widely used DSM-5 diagnostic criteria with analyses of mood course to deliver robust and explainable results. Extensive experiments validate the effectiveness of our method and the contribution of each design component. Future work will focus on pathways for real-world application, including integration with telehealth platforms and pilot studies within clinical settings to assess deployment efficacy and scalability.

\section*{Limitations}
This work's primary limitation is its reliance on a focused set of datasets. While we validated our approach on two different datasets, further validation across more diverse data sources is necessary to ensure broader generalizability. Furthermore, while our experimental setup simulates a realistic pre-screening scenario (e.g., by using a 1:19 imbalance ratio), validation in a real-world clinical setting is the gold standard and a necessary next step for deployment. Finally, future research should extend this approach to other mental health conditions, such as anxiety and bipolar disorder, to broaden its clinical relevance and impact.

\section*{Ethics Statement}

Our work on depression detection, a sensitive task, was guided by a strong ethical framework. We detail our primary considerations and mitigation strategies below.

\paragraph{Research Positioning and Intended Use.}
We position this work primarily as a Text Mining study aimed at extracting signals indicative of depression from unstructured text, aligning with established clinical criteria (DSM-5). We emphasize that this system is designed strictly as a \textit{supportive tool} for qualified mental health professionals. It is not an autonomous diagnostic system and should never be deployed directly to end-users or used to make final clinical decisions. The explainable evidence provided (e.g., symptom annotations) is intended to help professionals verify risk signals. This "human-in-the-loop" model is key to mitigating the risks of potential model bias and error inherent in LLMs. Any real-world implementation must be supervised by practitioners and preceded by thorough validation from clinical experts.

\paragraph{Data Privacy and Security.}
All data was sourced from users who provided explicit consent for their content to be used in mental health research. To protect user privacy, we moderately obfuscated all examples in this paper per established guidelines~\citep{bruckman2002studying}. While our research utilized a commercial API, we have engineered the system to be deployable with secure, locally-hosted open-source LLMs, thereby eliminating the need to share sensitive patient data with third-party services in a clinical setting.

\paragraph{Risks of Misuse and Mitigation.}
We recognize the potential for misuse, such as exploiting emotionally vulnerable individuals or causing stigmatization. To mitigate these risks, we recommend strict access controls, limiting system use to licensed mental health professionals. Clear documentation of the system's capabilities and limitations is crucial to ensure it is used responsibly as a supportive aid.

\paragraph{Broader Societal Impact.}
While our system aims to augment mental healthcare, we acknowledge that technology alone cannot solve systemic healthcare disparities. It should be considered one component within broader initiatives to expand mental health resources. We also note that this paper contains descriptions of depressive symptoms that may be distressing for some readers. We are committed to the ongoing ethical evaluation of our system and encourage community discussion on the responsible deployment of such technologies. Further discussion on the practical design considerations for deployment can be found in Appendix~\ref{sec:practical_design}.

\section*{Acknowledgement}
This work was supported in part by the National Key Research and Development Program of China under grant 2022YFB3104702, in part by the China National Natural Science Foundation of China under grant 72442026, 72342032 and 62272262.

\bibliography{references}

\begin{thebibliography}{77}
\providecommand{\natexlab}[1]{#1}

\bibitem[{Achiam et~al.(2023)Achiam, Adler, Agarwal, Ahmad, Akkaya, Aleman, Almeida, Altenschmidt, Altman, Anadkat et~al.}]{achiam2023gpt}
Josh Achiam, Steven Adler, Sandhini Agarwal, Lama Ahmad, Ilge Akkaya, Florencia~Leoni Aleman, Diogo Almeida, Janko Altenschmidt, Sam Altman, Shyamal Anadkat, and 1 others. 2023.
\newblock Gpt-4 technical report.
\newblock \emph{arXiv preprint arXiv:2303.08774}.

\bibitem[{Arcan et~al.(2024)Arcan, Niland, and Delahunty}]{arcan2024assessment}
Mihael Arcan, Paul-David Niland, and Fionn Delahunty. 2024.
\newblock An assessment on comprehending mental health through large language models.
\newblock \emph{arXiv preprint arXiv:2401.04592}.

\bibitem[{Balcombe and De~Leo(2021)}]{balcombe2021digital}
Luke Balcombe and Diego De~Leo. 2021.
\newblock Digital mental health challenges and the horizon ahead for solutions.
\newblock \emph{JMIR Mental Health}, 8(3):e26811.

\bibitem[{Blei et~al.(2003)Blei, Ng, and Jordan}]{blei2003latent}
David~M Blei, Andrew~Y Ng, and Michael~I Jordan. 2003.
\newblock Latent dirichlet allocation.
\newblock \emph{Journal of machine Learning research}, 3(Jan):993--1022.

\bibitem[{Brohan et~al.(2023)Brohan, Chebotar, Finn, Hausman, Herzog, Ho, Ibarz, Irpan, Jang, Julian et~al.}]{brohan2023can}
Anthony Brohan, Yevgen Chebotar, Chelsea Finn, Karol Hausman, Alexander Herzog, Daniel Ho, Julian Ibarz, Alex Irpan, Eric Jang, Ryan Julian, and 1 others. 2023.
\newblock Do as i can, not as i say: Grounding language in robotic affordances.
\newblock In \emph{Conference on robot learning}, pages 287--318. PMLR.

\bibitem[{Bruckman(2002)}]{bruckman2002studying}
Amy Bruckman. 2002.
\newblock Studying the amateur artist: A perspective on disguising data collected in human subjects research on the internet.
\newblock \emph{Ethics and Information Technology}, 4:217--231.

\bibitem[{Cacheda et~al.(2019)Cacheda, Fernandez, Novoa, and Carneiro}]{cacheda2019early}
Fidel Cacheda, Diego Fernandez, Francisco~J Novoa, and Victor Carneiro. 2019.
\newblock Early detection of depression: social network analysis and random forest techniques.
\newblock \emph{Journal of medical Internet research}, 21(6):e12554.

\bibitem[{Cai et~al.(2023)Cai, Wang, Ye, Jin, and Gao}]{cai2023depression}
Yicheng Cai, Haizhou Wang, Huali Ye, Yanwen Jin, and Wei Gao. 2023.
\newblock Depression detection on online social network with multivariate time series feature of user depressive symptoms.
\newblock \emph{Expert Systems with Applications}, 217:119538.

\bibitem[{Chen and Guestrin(2016)}]{chen2016xgboost}
Tianqi Chen and Carlos Guestrin. 2016.
\newblock Xgboost: A scalable tree boosting system.
\newblock In \emph{Proceedings of the 22nd acm sigkdd international conference on knowledge discovery and data mining}, pages 785--794.

\bibitem[{Chiong et~al.(2021)Chiong, Budhi, and Dhakal}]{chiong2021combining}
Raymond Chiong, Gregorious~Satia Budhi, and Sandeep Dhakal. 2021.
\newblock Combining sentiment lexicons and content-based features for depression detection.
\newblock \emph{IEEE Intelligent Systems}, 36(6):99--105.

\bibitem[{Cochran et~al.(2016)Cochran, McInnis, and Forger}]{cochran2016data}
AL~Cochran, MG~McInnis, and DB~Forger. 2016.
\newblock Data-driven classification of bipolar i disorder from longitudinal course of mood.
\newblock \emph{Translational psychiatry}, 6(10):e912--e912.

\bibitem[{Cohan et~al.(2018)Cohan, Desmet, Yates, Soldaini, MacAvaney, and Goharian}]{cohan2018smhd}
Arman Cohan, Bart Desmet, Andrew Yates, Luca Soldaini, Sean MacAvaney, and Nazli Goharian. 2018.
\newblock Smhd: a large-scale resource for exploring online language usage for multiple mental health conditions.
\newblock \emph{arXiv preprint arXiv:1806.05258}.

\bibitem[{Coppersmith et~al.(2014)Coppersmith, Dredze, and Harman}]{coppersmith2014quantifying}
Glen Coppersmith, Mark Dredze, and Craig Harman. 2014.
\newblock Quantifying mental health signals in twitter.
\newblock In \emph{Proceedings of the workshop on computational linguistics and clinical psychology: From linguistic signal to clinical reality}, pages 51--60.

\bibitem[{Costello et~al.(2002)Costello, Pine, Hammen, March, Plotsky, Weissman, Biederman, Goldsmith, Kaufman, Lewinsohn et~al.}]{costello2002development}
E~Jane Costello, Daniel~S Pine, Constance Hammen, John~S March, Paul~M Plotsky, Myrna~M Weissman, Joseph Biederman, H~Hill Goldsmith, Joan Kaufman, Peter~M Lewinsohn, and 1 others. 2002.
\newblock Development and natural history of mood disorders.
\newblock \emph{Biological psychiatry}, 52(6):529--542.

\bibitem[{Davis and Goadrich(2006)}]{davis2006relationship}
Jesse Davis and Mark Goadrich. 2006.
\newblock The relationship between precision-recall and roc curves.
\newblock In \emph{Proceedings of the 23rd international conference on Machine learning}, pages 233--240.

\bibitem[{Devlin et~al.(2018)Devlin, Chang, Lee, and Toutanova}]{devlin2018bert}
Jacob Devlin, Ming-Wei Chang, Kenton Lee, and Kristina Toutanova. 2018.
\newblock Bert: Pre-training of deep bidirectional transformers for language understanding.
\newblock \emph{arXiv preprint arXiv:1810.04805}.

\bibitem[{Du et~al.(2024)Du, Feng, Zhao, Yuan, and Li}]{du2024trajagent}
Yuwei Du, Jie Feng, Jie Zhao, Jian Yuan, and Yong Li. 2024.
\newblock Trajagent: An llm-based agent framework for automated trajectory modeling via collaboration of large and small models.
\newblock \emph{arXiv preprint arXiv:2410.20445}.

\bibitem[{Eichstaedt et~al.(2018)Eichstaedt, Smith, Merchant, Ungar, Crutchley, Preo{\c{t}}iuc-Pietro, Asch, and Schwartz}]{eichstaedt2018facebook}
Johannes~C Eichstaedt, Robert~J Smith, Raina~M Merchant, Lyle~H Ungar, Patrick Crutchley, Daniel Preo{\c{t}}iuc-Pietro, David~A Asch, and H~Andrew Schwartz. 2018.
\newblock Facebook language predicts depression in medical records.
\newblock \emph{Proceedings of the National Academy of Sciences}, 115(44):11203--11208.

\bibitem[{Feng et~al.(2024)Feng, Du, Zhao, and Li}]{feng2024agentmove}
Jie Feng, Yuwei Du, Jie Zhao, and Yong Li. 2024.
\newblock Agentmove: A large language model based agentic framework for zero-shot next location prediction.
\newblock \emph{arXiv preprint arXiv:2408.13986}.

\bibitem[{Fr{\'\i}as et~al.(2017)Fr{\'\i}as, Dickstein, Merranko, Gill, Goldstein, Goldstein, Hower, Yen, Hafeman, Liao et~al.}]{frias2017longitudinal}
{\'A}lvaro Fr{\'\i}as, Daniel~P Dickstein, John Merranko, Mary~Kay Gill, Tina~R Goldstein, Benjamin~I Goldstein, Heather Hower, Shirley Yen, Danella~M Hafeman, Fangzi Liao, and 1 others. 2017.
\newblock Longitudinal cognitive trajectories and associated clinical variables in youth with bipolar disorder.
\newblock \emph{Bipolar disorders}, 19(4):273--284.

\bibitem[{Gao et~al.(2024)Gao, Lan, Li, Yuan, Ding, Zhou, Xu, and Li}]{gao2024large}
Chen Gao, Xiaochong Lan, Nian Li, Yuan Yuan, Jingtao Ding, Zhilun Zhou, Fengli Xu, and Yong Li. 2024.
\newblock Large language models empowered agent-based modeling and simulation: A survey and perspectives.
\newblock \emph{Humanities and Social Sciences Communications}, 11(1):1--24.

\bibitem[{Gao et~al.(2023)Gao, Lan, Lu, Mao, Piao, Wang, Jin, and Li}]{gao2023s3}
Chen Gao, Xiaochong Lan, Zhihong Lu, Jinzhu Mao, Jinghua Piao, Huandong Wang, Depeng Jin, and Yong Li. 2023.
\newblock S3: Social-network simulation system with large language model-empowered agents.
\newblock \emph{arXiv preprint arXiv:2307.14984}.

\bibitem[{Hao et~al.(2024)Hao, Fan, Xu, Yuan, and Li}]{hao2024hlm}
Qianyue Hao, Jingyang Fan, Fengli Xu, Jian Yuan, and Yong Li. 2024.
\newblock Hlm-cite: Hybrid language model workflow for text-based scientific citation prediction.
\newblock \emph{Advances in Neural Information Processing Systems}, 37:48189--48223.

\bibitem[{Hao et~al.(2025{\natexlab{a}})Hao, Li, Yuan, and Li}]{hao2025rl}
Qianyue Hao, Sibo Li, Jian Yuan, and Yong Li. 2025{\natexlab{a}}.
\newblock Rl of thoughts: Navigating llm reasoning with inference-time reinforcement learning.
\newblock \emph{arXiv preprint arXiv:2505.14140}.

\bibitem[{HAO et~al.(2024)HAO, QI, YUAN, ZONG, CHEN, ZHAO, WANG, ZHANG, YUAN, and LI}]{hao2024reinforcement}
QIANYUE HAO, XIAOQIAN QI, YUAN YUAN, ZEFANG ZONG, HONGYI CHEN, KEYU ZHAO, SHENGYUAN WANG, YUNKE ZHANG, JIAN YUAN, and YONG LI. 2024.
\newblock Reinforcement learning in the era of large language models: Challenges and opportunities.

\bibitem[{Hao et~al.(2025{\natexlab{b}})Hao, Song, Liao, Yuan, and Li}]{hao2025llm}
Qianyue Hao, Yiwen Song, Qingmin Liao, Jian Yuan, and Yong Li. 2025{\natexlab{b}}.
\newblock Llm-explorer: A plug-in reinforcement learning policy exploration enhancement driven by large language models.
\newblock \emph{arXiv preprint arXiv:2505.15293}.

\bibitem[{Hou et~al.(2024)Hou, Zhao, Liu, Yang, Wang, Li, Luo, Lo, Grundy, and Wang}]{hou2024large}
Xinyi Hou, Yanjie Zhao, Yue Liu, Zhou Yang, Kailong Wang, Li~Li, Xiapu Luo, David Lo, John Grundy, and Haoyu Wang. 2024.
\newblock Large language models for software engineering: A systematic literature review.
\newblock \emph{ACM Transactions on Software Engineering and Methodology}, 33(8):1--79.

\bibitem[{Hua et~al.(2024)Hua, Liu, Yang, Li, Sheu, Zhou, Moran, Ananiadou, and Beam}]{hua2024large}
Yining Hua, Fenglin Liu, Kailai Yang, Zehan Li, Yi-han Sheu, Peilin Zhou, Lauren~V Moran, Sophia Ananiadou, and Andrew Beam. 2024.
\newblock Large language models in mental health care: a scoping review.
\newblock \emph{arXiv preprint arXiv:2401.02984}.

\bibitem[{Ive et~al.(2018)Ive, Gkotsis, Dutta, Stewart, and Velupillai}]{ive2018hierarchical}
Julia Ive, George Gkotsis, Rina Dutta, Robert Stewart, and Sumithra Velupillai. 2018.
\newblock Hierarchical neural model with attention mechanisms for the classification of social media text related to mental health.
\newblock In \emph{Proceedings of the fifth workshop on computational linguistics and clinical psychology: from keyboard to clinic}, pages 69--77.

\bibitem[{Ji et~al.(2023)Ji, Zhang, Yang, Ananiadou, Cambria, and Tiedemann}]{ji2023domain}
Shaoxiong Ji, Tianlin Zhang, Kailai Yang, Sophia Ananiadou, Erik Cambria, and J{\"o}rg Tiedemann. 2023.
\newblock Domain-specific continued pretraining of language models for capturing long context in mental health.
\newblock \emph{arXiv preprint arXiv:2304.10447}.

\bibitem[{Jimenez et~al.(2023)Jimenez, Yang, Wettig, Yao, Pei, Press, and Narasimhan}]{jimenez2023swe}
Carlos~E Jimenez, John Yang, Alexander Wettig, Shunyu Yao, Kexin Pei, Ofir Press, and Karthik Narasimhan. 2023.
\newblock Swe-bench: Can language models resolve real-world github issues?
\newblock \emph{arXiv preprint arXiv:2310.06770}.

\bibitem[{Joulin et~al.(2016)Joulin, Grave, Bojanowski, Douze, J{\'e}gou, and Mikolov}]{joulin2016fasttext}
Armand Joulin, Edouard Grave, Piotr Bojanowski, Matthijs Douze, H{\'e}rve J{\'e}gou, and Tomas Mikolov. 2016.
\newblock Fasttext. zip: Compressing text classification models.
\newblock \emph{arXiv preprint arXiv:1612.03651}.

\bibitem[{Khan et~al.(2016)}]{khan2016economic}
Murad~Moosa Khan and 1 others. 2016.
\newblock Economic burden of mental illnesses in pakistan.
\newblock \emph{Journal of Mental Health Policy and Economics}, 19(3):155.

\bibitem[{Lan et~al.(2024{\natexlab{a}})Lan, Cheng, Sheng, Gao, and Li}]{lan2024depression}
Xiaochong Lan, Yiming Cheng, Li~Sheng, Chen Gao, and Yong Li. 2024{\natexlab{a}}.
\newblock Depression detection on social media with large language models.
\newblock \emph{arXiv preprint arXiv:2403.10750}.

\bibitem[{Lan et~al.(2025{\natexlab{a}})Lan, Feng, Lei, Shi, and Li}]{lan2025benchmarking}
Xiaochong Lan, Jie Feng, Jiahuan Lei, Xinlei Shi, and Yong Li. 2025{\natexlab{a}}.
\newblock Benchmarking and advancing large language models for local life services.
\newblock In \emph{Proceedings of the 31st ACM SIGKDD Conference on Knowledge Discovery and Data Mining V. 2}, pages 4566--4577.

\bibitem[{Lan et~al.(2025{\natexlab{b}})Lan, Feng, Sun, Gao, Lei, Shi, Luo, and Li}]{lan2025open}
Xiaochong Lan, Jie Feng, Yizhou Sun, Chen Gao, Jiahuan Lei, Xinlei Shi, Hengliang Luo, and Yong Li. 2025{\natexlab{b}}.
\newblock Open-set living need prediction with large language models.
\newblock \emph{arXiv preprint arXiv:2506.02713}.

\bibitem[{Lan et~al.(2024{\natexlab{b}})Lan, Gao, Jin, and Li}]{lan2024stance}
Xiaochong Lan, Chen Gao, Depeng Jin, and Yong Li. 2024{\natexlab{b}}.
\newblock Stance detection with collaborative role-infused llm-based agents.
\newblock In \emph{Proceedings of the international AAAI conference on web and social media}, volume~18, pages 891--903.

\bibitem[{Li et~al.(2023)Li, Zhang, Zhang, Long, Xie, and Zhang}]{li2023towards}
Zehan Li, Xin Zhang, Yanzhao Zhang, Dingkun Long, Pengjun Xie, and Meishan Zhang. 2023.
\newblock Towards general text embeddings with multi-stage contrastive learning.
\newblock \emph{arXiv preprint arXiv:2308.03281}.

\bibitem[{Lin et~al.(2020)Lin, Hu, Su, Li, Mei, Zhou, and Leung}]{lin2020sensemood}
Chenhao Lin, Pengwei Hu, Hui Su, Shaochun Li, Jing Mei, Jie Zhou, and Henry Leung. 2020.
\newblock Sensemood: depression detection on social media.
\newblock In \emph{Proceedings of the 2020 international conference on multimedia retrieval}, pages 407--411.

\bibitem[{Losada and Crestani(2016)}]{losada2016test}
David~E Losada and Fabio Crestani. 2016.
\newblock A test collection for research on depression and language use.
\newblock In \emph{International conference of the cross-language evaluation forum for European languages}, pages 28--39. Springer.

\bibitem[{Malhotra and Jindal(2022)}]{malhotra2022deep}
Anshu Malhotra and Rajni Jindal. 2022.
\newblock Deep learning techniques for suicide and depression detection from online social media: A scoping review.
\newblock \emph{Applied Soft Computing}, page 109713.

\bibitem[{Mon-Williams et~al.(2025)Mon-Williams, Li, Long, Du, and Lucas}]{mon2025embodied}
Ruaridh Mon-Williams, Gen Li, Ran Long, Wenqian Du, and Christopher~G Lucas. 2025.
\newblock Embodied large language models enable robots to complete complex tasks in unpredictable environments.
\newblock \emph{Nature Machine Intelligence}, pages 1--10.

\bibitem[{Naseem et~al.(2022)Naseem, Dunn, Kim, and Khushi}]{naseem2022early}
Usman Naseem, Adam~G Dunn, Jinman Kim, and Matloob Khushi. 2022.
\newblock Early identification of depression severity levels on reddit using ordinal classification.
\newblock In \emph{Proceedings of the ACM Web Conference 2022}, pages 2563--2572.

\bibitem[{Nguyen et~al.(2022)Nguyen, Yates, Zirikly, Desmet, and Cohan}]{nguyen2022improving}
Thong Nguyen, Andrew Yates, Ayah Zirikly, Bart Desmet, and Arman Cohan. 2022.
\newblock Improving the generalizability of depression detection by leveraging clinical questionnaires.
\newblock \emph{arXiv preprint arXiv:2204.10432}.

\bibitem[{Oatley and Johnson-Laird(1987)}]{oatley1987towards}
Keith Oatley and Philip~N Johnson-Laird. 1987.
\newblock Towards a cognitive theory of emotions.
\newblock \emph{Cognition and emotion}, 1(1):29--50.

\bibitem[{Orabi et~al.(2018)Orabi, Buddhitha, Orabi, and Inkpen}]{orabi2018deep}
Ahmed~Husseini Orabi, Prasadith Buddhitha, Mahmoud~Husseini Orabi, and Diana Inkpen. 2018.
\newblock Deep learning for depression detection of twitter users.
\newblock In \emph{Proceedings of the fifth workshop on computational linguistics and clinical psychology: from keyboard to clinic}, pages 88--97.

\bibitem[{Organization()}]{worldint}
World~Health Organization.
\newblock www. who. int/news-room/fact-sheets/detail/depression.

\bibitem[{Ouyang et~al.(2022)Ouyang, Wu, Jiang, Almeida, Wainwright, Mishkin, Zhang, Agarwal, Slama, Ray et~al.}]{ouyang2022training}
Long Ouyang, Jeffrey Wu, Xu~Jiang, Diogo Almeida, Carroll Wainwright, Pamela Mishkin, Chong Zhang, Sandhini Agarwal, Katarina Slama, Alex Ray, and 1 others. 2022.
\newblock Training language models to follow instructions with human feedback.
\newblock \emph{Advances in neural information processing systems}, 35:27730--27744.

\bibitem[{Owen et~al.(2020)Owen, Collados, and Espinosa-Anke}]{owen2020towards}
David Owen, Jose~Camacho Collados, and Luis Espinosa-Anke. 2020.
\newblock Towards preemptive detection of depression and anxiety in twitter.
\newblock \emph{arXiv preprint arXiv:2011.05249}.

\bibitem[{Piao et~al.(2025)Piao, Yan, Zhang, Li, Yan, Lan, Lu, Zheng, Wang, Zhou et~al.}]{piao2025agentsociety}
Jinghua Piao, Yuwei Yan, Jun Zhang, Nian Li, Junbo Yan, Xiaochong Lan, Zhihong Lu, Zhiheng Zheng, Jing~Yi Wang, Di~Zhou, and 1 others. 2025.
\newblock Agentsociety: Large-scale simulation of llm-driven generative agents advances understanding of human behaviors and society.
\newblock \emph{arXiv preprint arXiv:2502.08691}.

\bibitem[{Ramos et~al.(2003)}]{ramos2003using}
Juan Ramos and 1 others. 2003.
\newblock Using tf-idf to determine word relevance in document queries.
\newblock In \emph{Proceedings of the first instructional conference on machine learning}, volume 242, pages 29--48. Citeseer.

\bibitem[{Regier et~al.(2013)Regier, Kuhl, and Kupfer}]{regier2013dsm}
Darrel~A Regier, Emily~A Kuhl, and David~J Kupfer. 2013.
\newblock The dsm-5: Classification and criteria changes.
\newblock \emph{World psychiatry}, 12(2):92--98.

\bibitem[{Ross et~al.(2023)Ross, Martinez, Houde, Muller, and Weisz}]{ross2023programmer}
Steven~I Ross, Fernando Martinez, Stephanie Houde, Michael Muller, and Justin~D Weisz. 2023.
\newblock The programmer’s assistant: Conversational interaction with a large language model for software development.
\newblock In \emph{Proceedings of the 28th International Conference on Intelligent User Interfaces}, pages 491--514.

\bibitem[{Rudolph et~al.(2006)Rudolph, Hammen, and Daley}]{rudolph2006mood}
Karen~D Rudolph, Constance Hammen, and Shannon~E Daley. 2006.
\newblock Mood disorders.
\newblock \emph{Behavioral and emotional disorders in adolescents: Nature, assessment, and treatment}, pages 300--342.

\bibitem[{R{\"u}sch et~al.(2005)R{\"u}sch, Angermeyer, and Corrigan}]{rusch2005mental}
Nicolas R{\"u}sch, Matthias~C Angermeyer, and Patrick~W Corrigan. 2005.
\newblock Mental illness stigma: Concepts, consequences, and initiatives to reduce stigma.
\newblock \emph{European psychiatry}, 20(8):529--539.

\bibitem[{Sarkar et~al.(2023)Sarkar, Alhamadani, Behal, Alkulaib, and Lu}]{sarkar2023analyzing}
Sarkar Sarkar, Abdulaziz Alhamadani, Sristhi Behal, Lulwah Alkulaib, and Chang-Tien Lu. 2023.
\newblock Analyzing prediction of depression and anxiety on reddit: a multi-task learning approach through gmmtl.

\bibitem[{Sarkar et~al.(2022)Sarkar, Alhamadani, Alkulaib, and Lu}]{sarkar2022predicting}
Shailik Sarkar, Abdulaziz Alhamadani, Lulwah Alkulaib, and Chang-Tien Lu. 2022.
\newblock Predicting depression and anxiety on reddit: a multi-task learning approach.
\newblock In \emph{2022 IEEE/ACM International Conference on Advances in Social Networks Analysis and Mining (ASONAM)}, pages 427--435. IEEE.

\bibitem[{Shen et~al.(2017)Shen, Jia, Nie, Feng, Zhang, Hu, Chua, Zhu et~al.}]{shen2017depression}
Guangyao Shen, Jia Jia, Liqiang Nie, Fuli Feng, Cunjun Zhang, Tianrui Hu, Tat-Seng Chua, Wenwu Zhu, and 1 others. 2017.
\newblock Depression detection via harvesting social media: A multimodal dictionary learning solution.

\bibitem[{Singhal et~al.(2023)Singhal, Azizi, Tu, Mahdavi, Wei, Chung, Scales, Tanwani, Cole-Lewis, Pfohl et~al.}]{singhal2023large}
Karan Singhal, Shekoofeh Azizi, Tao Tu, S~Sara Mahdavi, Jason Wei, Hyung~Won Chung, Nathan Scales, Ajay Tanwani, Heather Cole-Lewis, Stephen Pfohl, and 1 others. 2023.
\newblock Large language models encode clinical knowledge.
\newblock \emph{Nature}, 620(7972):172--180.

\bibitem[{Tadesse et~al.(2019)Tadesse, Lin, Xu, and Yang}]{tadesse2019detection}
Michael~M Tadesse, Hongfei Lin, Bo~Xu, and Liang Yang. 2019.
\newblock Detection of depression-related posts in reddit social media forum.
\newblock \emph{Ieee Access}, 7:44883--44893.

\bibitem[{Tausczik and Pennebaker(2010)}]{tausczik2010psychological}
Yla~R Tausczik and James~W Pennebaker. 2010.
\newblock The psychological meaning of words: Liwc and computerized text analysis methods.
\newblock \emph{Journal of language and social psychology}, 29(1):24--54.

\bibitem[{Touvron et~al.(2023)Touvron, Martin, Stone, Albert, Almahairi, Babaei, Bashlykov, Batra, Bhargava, Bhosale et~al.}]{touvron2023llama}
Hugo Touvron, Louis Martin, Kevin Stone, Peter Albert, Amjad Almahairi, Yasmine Babaei, Nikolay Bashlykov, Soumya Batra, Prajjwal Bhargava, Shruti Bhosale, and 1 others. 2023.
\newblock Llama 2: Open foundation and fine-tuned chat models.
\newblock \emph{arXiv preprint arXiv:2307.09288}.

\bibitem[{Wang et~al.(2025)Wang, Shi, Hu, Ma, Liu, Wang, Yao, Liu, Ge, and Zhang}]{wang2025large}
Jiaqi Wang, Enze Shi, Huawen Hu, Chong Ma, Yiheng Liu, Xuhui Wang, Yincheng Yao, Xuan Liu, Bao Ge, and Shu Zhang. 2025.
\newblock Large language models for robotics: Opportunities, challenges, and perspectives.
\newblock \emph{Journal of Automation and Intelligence}, 4(1):52--64.

\bibitem[{Whitton et~al.(2021)Whitton, Hardy, Cope, Gieng, Gow, MacKinnon, Gale, O'Moore, Anderson, Proudfoot et~al.}]{whitton2021mental}
Alexis~E Whitton, Rebecca Hardy, Kate Cope, Chilin Gieng, Leanne Gow, Andrew MacKinnon, Nyree Gale, Kathleen O'Moore, Josephine Anderson, Judith Proudfoot, and 1 others. 2021.
\newblock Mental health screening in general practices as a means for enhancing uptake of digital mental health interventions: observational cohort study.
\newblock \emph{Journal of medical Internet research}, 23(9):e28369.

\bibitem[{Wu et~al.(2023)Wu, Wu, Hua, Lin, Zheng, and Yang}]{wu2023exploring}
Jiageng Wu, Xian Wu, Yining Hua, Shixu Lin, Yefeng Zheng, and Jie Yang. 2023.
\newblock Exploring social media for early detection of depression in covid-19 patients.
\newblock \emph{arXiv preprint arXiv:2302.12044}.

\bibitem[{Xu et~al.(2025)Xu, Hao, Zong, Wang, Zhang, Wang, Lan, Gong, Ouyang, Meng et~al.}]{xu2025towards}
Fengli Xu, Qianyue Hao, Zefang Zong, Jingwei Wang, Yunke Zhang, Jingyi Wang, Xiaochong Lan, Jiahui Gong, Tianjian Ouyang, Fanjin Meng, and 1 others. 2025.
\newblock Towards large reasoning models: A survey of reinforced reasoning with large language models.
\newblock \emph{arXiv preprint arXiv:2501.09686}.

\bibitem[{Yadav et~al.(2020)Yadav, Chauhan, Sain, Thirunarayan, Sheth, and Schumm}]{yadav2020identifying}
Shweta Yadav, Jainish Chauhan, Joy~Prakash Sain, Krishnaprasad Thirunarayan, Amit Sheth, and Jeremiah Schumm. 2020.
\newblock Identifying depressive symptoms from tweets: Figurative language enabled multitask learning framework.
\newblock \emph{arXiv preprint arXiv:2011.06149}.

\bibitem[{Yan et~al.(2025)Yan, Piao, Lan, Shao, Hui, and Li}]{yan2025simulating}
Yuwei Yan, Jinghua Piao, Xiaochong Lan, Chenyang Shao, Pan Hui, and Yong Li. 2025.
\newblock Simulating generative social agents via theory-informed workflow design.
\newblock \emph{arXiv preprint arXiv:2508.08726}.

\bibitem[{Yang et~al.(2023{\natexlab{a}})Yang, Ji, Zhang, Xie, Kuang, and Ananiadou}]{yang2023towards}
Kailai Yang, Shaoxiong Ji, Tianlin Zhang, Qianqian Xie, Ziyan Kuang, and Sophia Ananiadou. 2023{\natexlab{a}}.
\newblock Towards interpretable mental health analysis with large language models.
\newblock In \emph{Proceedings of the 2023 Conference on Empirical Methods in Natural Language Processing}, pages 6056--6077.

\bibitem[{Yang et~al.(2022)Yang, Zhang, and Ananiadou}]{yang2022mental}
Kailai Yang, Tianlin Zhang, and Sophia Ananiadou. 2022.
\newblock A mental state knowledge--aware and contrastive network for early stress and depression detection on social media.
\newblock \emph{Information Processing \& Management}, 59(4):102961.

\bibitem[{Yang et~al.(2023{\natexlab{b}})Yang, Zhang, Kuang, Xie, and Ananiadou}]{yang2023mentalllama}
Kailai Yang, Tianlin Zhang, Ziyan Kuang, Qianqian Xie, and Sophia Ananiadou. 2023{\natexlab{b}}.
\newblock Mentalllama: Interpretable mental health analysis on social media with large language models.
\newblock \emph{arXiv preprint arXiv:2309.13567}.

\bibitem[{Yang et~al.(2016)Yang, Yang, Dyer, He, Smola, and Hovy}]{yang2016hierarchical}
Zichao Yang, Diyi Yang, Chris Dyer, Xiaodong He, Alex Smola, and Eduard Hovy. 2016.
\newblock Hierarchical attention networks for document classification.
\newblock In \emph{Proceedings of the 2016 conference of the North American chapter of the association for computational linguistics: human language technologies}, pages 1480--1489.

\bibitem[{Zhang et~al.(2022)Zhang, Schoene, Ji, and Ananiadou}]{zhang2022natural}
Tianlin Zhang, Annika~M Schoene, Shaoxiong Ji, and Sophia Ananiadou. 2022.
\newblock Natural language processing applied to mental illness detection: a narrative review.
\newblock \emph{NPJ digital medicine}, 5(1):46.

\bibitem[{Zhang et~al.(2023{\natexlab{a}})Zhang, Yang, Alhuzali, Liu, and Ananiadou}]{zhang2023phq}
Tianlin Zhang, Kailai Yang, Hassan Alhuzali, Boyang Liu, and Sophia Ananiadou. 2023{\natexlab{a}}.
\newblock Phq-aware depressive symptoms identification with similarity contrastive learning on social media.
\newblock \emph{Information Processing \& Management}, 60(5):103417.

\bibitem[{Zhang et~al.(2023{\natexlab{b}})Zhang, Yang, and Ananiadou}]{zhang2023sentiment}
Tianlin Zhang, Kailai Yang, and Sophia Ananiadou. 2023{\natexlab{b}}.
\newblock Sentiment-guided transformer with severity-aware contrastive learning for depression detection on social media.
\newblock In \emph{The 22nd Workshop on Biomedical Natural Language Processing and BioNLP Shared Tasks}, pages 114--126.

\bibitem[{Ziems et~al.(2023)Ziems, Held, Shaikh, Chen, Zhang, and Yang}]{ziems2023can}
Caleb Ziems, William Held, Omar Shaikh, Jiaao Chen, Zhehao Zhang, and Diyi Yang. 2023.
\newblock Can large language models transform computational social science?
\newblock \emph{arXiv preprint arXiv:2305.03514}.

\bibitem[{Zogan et~al.(2021)Zogan, Razzak, Jameel, and Xu}]{zogan2021depressionnet}
Hamad Zogan, Imran Razzak, Shoaib Jameel, and Guandong Xu. 2021.
\newblock Depressionnet: A novel summarization boosted deep framework for depression detection on social media.
\newblock \emph{arXiv preprint arXiv:2105.10878}.

\end{thebibliography}

\clearpage

\appendix

\section{Related Works}
\subsection{Depression Detection on Social Media}
Compared to traditional medical diagnostic methods for depression conducted in hospitals, depression detection on social media has the advantages of lower concealment potential, wider coverage, and lower cost~\citep{malhotra2022deep}. Early research works first extract text features, then apply traditional machine learning methods for classification. Common feature extraction methods in depression detection research include LIWC~\citep{tausczik2010psychological}, TF-IDF, LDA~\citep{blei2003latent}, etc.; classifiers include SVM~\citep{tadesse2019detection}, Logistic Regression~\citep{coppersmith2014quantifying}, Random Forest~\citep{cacheda2019early}, among others. 

In recent years, deep learning technologies have seen wide application in depression detection. Methods utilizing CNN~\citep{cohan2018smhd} and RNN~\citep{ive2018hierarchical} have been developed to capture complex patterns in text. Pre-trained Language Models (PLMs)~\citep{owen2020towards,ji2023domain} have further improved performance by leveraging knowledge learned from large corpora. Graph Convolutional Networks (GCNs)~\citep{naseem2022early} have also been employed to model user interactions and behaviors. Notably, AMM-Net~\citep{sarkar2022predicting} introduced an a multi-task learning model that combines word embeddings and topic features to effectively predict depression and anxiety on Reddit. Follow-up work extended this approach using Graph-based Message-passing Multi-task learning (GMMTL)~\citep{sarkar2023analyzing}. These efforts have significantly improved the accuracy of depression detection, advancing the field's research.

Some works on depression detection focus solely on analyzing individual texts to determine the presence of depressive symptoms~\citep{yang2022mental,chiong2021combining}, the type of symptoms displayed~\citep{zhang2023phq}, and the level of depression~\citep{zhang2023sentiment}. However, analyzing a user's post history is more informative, as it is common for individuals not suffering from depression to also occasionally post texts that exhibit depressive symptoms. Recent approaches to depression detection based on a user's post history~\cite{eichstaedt2018facebook,orabi2018deep,lin2020sensemood,chiong2021combining,yang2022mental,sarkar2022predicting} lack comprehensive integration of medical knowledge, limiting their accuracy and explainability. While some studies have attempted to introduce medical knowledge by considering clinical symptoms~\citep{yadav2020identifying,nguyen2022improving}, they do so through relatively ineffective methods such as keyword matching or vector similarity matching, which may miss complex expressions of symptoms. Our research stands out as one of the initial efforts to more systematically apply medical insights, leading to improved accuracy. Additionally, we are pioneers in employing LLMs to craft systematic explanations of predictions grounded in clinical evidence, thereby increasing explainability.
\subsection{Large Language Models}
Large Language Models (LLMs), possessing rich common-sense knowledge and human-like reasoning abilities, have achieved tremendous success recently. LLMs are being applied to a wide variety of domains, such as mathematics and reasoning~\cite{xu2025towards,hao2025llm, hao2025rl, hao2024reinforcement}, text analysis~\cite{hao2024hlm,lan2024stance, lan2024depression}, user behavior modeling~\cite{lan2025benchmarking,lan2025open,feng2024agentmove,du2024trajagent}, user simulation~\cite{gao2023s3,piao2025agentsociety,gao2024large,yan2025simulating}, code generation~\cite{jimenez2023swe,hou2024large,ross2023programmer}, and embodied intelligence~\cite{brohan2023can,wang2025large,mon2025embodied}. Trained on vast amounts of human-generated text, LLMs have a strong potential to understand people, making them suitable for tasks like user behavior modeling and text comprehension. Consequently, they are also likely well-suited for tasks that require a deep understanding of human inner states, such as mental health analysis. Our work is among the first to explore this direction.

\subsection{LLMs for Mental Health Analysis}
Large language models (LLMs) have shown great potential in clinical applications due to its abundant prior knowledge and strong language generalization ability~\citep{singhal2023large}. Recent works have introduced LLMs into mental health analysis. \cite{arcan2024assessment} presents a solid evaluation on the performance of Llama-2~\citep{touvron2023llama} and ChatGPT~\cite{ouyang2022training} in mental health assessment tasks, unveiling prospects along with challenges of LLM-based methods. \citet{yang2023towards} further discover the applications of LLMs in both mental health detection tasks and reasoning tasks, which highlight LLM's excellent interpretability. \citet{yang2023mentalllama} propose the first open-source LLM series for mental health analysis based on Llama-2 and fine-tuning techniques, and greatly enhances the accuracy and explanation quality compared with general-purpose LLMs. However, the challenge of LLM-based methods being weaker in terms of prediction accuracy compared to embedding model-based methods, especially when targeting specific downstream tasks, remains unresolved~\citep{arcan2024assessment}.

To our knowledge, our system is among the first to be specifically designed for depression detection in the context of LLM for mental health analysis. Furthermore, we have facilitated a collaboration between LLMs and embedding models, attaining both high accuracy and explainability, addressing deficiencies in previous designs.
\section{Datasets}
\label{sec:dataset}

\subsection{SWDD Dataset}
Acquiring well-labeled, large-scale datasets for depression detection is challenging due to the sensitivity and privacy concerns associated with mental illnesses. Some datasets utilize self-reported medical diagnoses. While this approach offers relatively high reliability, it leads to datasets comprising users who are already medically diagnosed and willing to discuss it online. However, the most valuable aspect of the depression detection task lies in assisting human experts to identify potentially undiagnosed users. Furthermore, the ratio of depressed to normal users in most previous datasets~\cite{losada2016test,zogan2021depressionnet,wu2023exploring} is around 1:4 or even 1:1, deviating far from the actual prevalence of about 5\% in the population.

To address these issues, we utilize the SWDD dataset~\cite{cai2023depression}. This dataset carefully adheres to strict ethical guidelines. It includes users who self-disclosed a diagnosis and those identified by experts as highly likely to be depressed based on medical criteria, but without self-disclosure. We retain these non-self-disclosing users and, after further expert screening, keep 1000 depressed users and 19000 normal users to simulate a real-world application scenario with a 1:19 ratio. We retain only the posts posted within six months prior to each user's last post. The dataset statistics are shown in Table~\ref{tab:dataset}. Training, validation, and test sets were divided in a 7:1:2 ratio, as recommended by~\cite{wu2023exploring}.

\subsection{Twitter Mental Disorder Dataset}
To further validate the generalizability of our approach, we also utilized the Twitter Mental Disorder Tweets and Musics dataset. This dataset contains tweets from users diagnosed with various mental disorders, including depression, alongside a control group. We utilized the subset relevant to depression detection for our experiments.

\begin{table}[h!]
\centering
\caption{Statistics of the SWDD dataset.}
\label{tab:dataset}
\resizebox{\columnwidth}{!}{%
\begin{tabular}{lrr}
\hline
& \textbf{Depressed} & \textbf{Control} \\
\hline
\textbf{Num. of users} & 1,000 & 19,000 \\
\textbf{Num. of posts} & 69,548 & 1,314,874 \\
\textbf{Avg. num. of posts per user} & 69.55 & 69.20 \\
\hline
\end{tabular}%
}
\end{table}

\section{Details of Comparison Methods}
\label{sec:baseline}
Here, we give a more detailed discussion of our utilized comparison methods.
\begin{itemize}[leftmargin=*]
\item TF-IDF+XGBoost~\citep{ramos2003using,chen2016xgboost,wu2023exploring}: It uses TF-IDF weighted features of word and character n-grams for feature extraction, followed by XGBoost for classification.
\item HAN~\citep{yang2016hierarchical}: It obtains tweet-level representations through bidirectional GRU networks, and encodes all representations into user presentation with an attention mechanism.
\item Mood2Content~\citep{cai2023depression}: This method utilizes knowledge distillation technology, transferring knowledge from an emotion classifier to depression detection.
\item AMM-Net~\citep{sarkar2022predicting}: An attention-based multi-view and multi-task learning framework that integrates different feature perspectives and jointly optimizes for related mental health conditions (e.g., depression and anxiety).
\item FastText~\citep{joulin2016fasttext}, gte-small~\cite{li2023towards}, BERT~\citep{devlin2018bert}, MentalRoBERTa~\citep{ji2023domain}: These models are used to obtain text embeddings. We compute the average embedding across all posts to generate a user-level representation, which is then fed into a tree classifier.
\item PHQ9 (Score) and PHQ9 (Vector)~\cite{nguyen2022improving}: PHQ9 (Score) uses BERT to classify the 9 symptoms of PHQ-9, and then uses these symptom scores as features input to a CNN classifier. PHQ9 (Vector) uses the hidden layer vectors of the PHQ-9 symptom classifiers as features.
\item GPT-4o mini, MentalLLaMA~\citep{yang2023mentalllama}: For LLM-based approaches, we design prompts following~\citet{yang2023towards}. We input the user's tweet history to the LLM, and then ask the LLM to classify.
\end{itemize}

\section{Hyperparameter Study (RQ4)}
\begin{figure}[]
\centering
\subfigure[Impact of $k$.]{
\includegraphics[width=2.4cm]{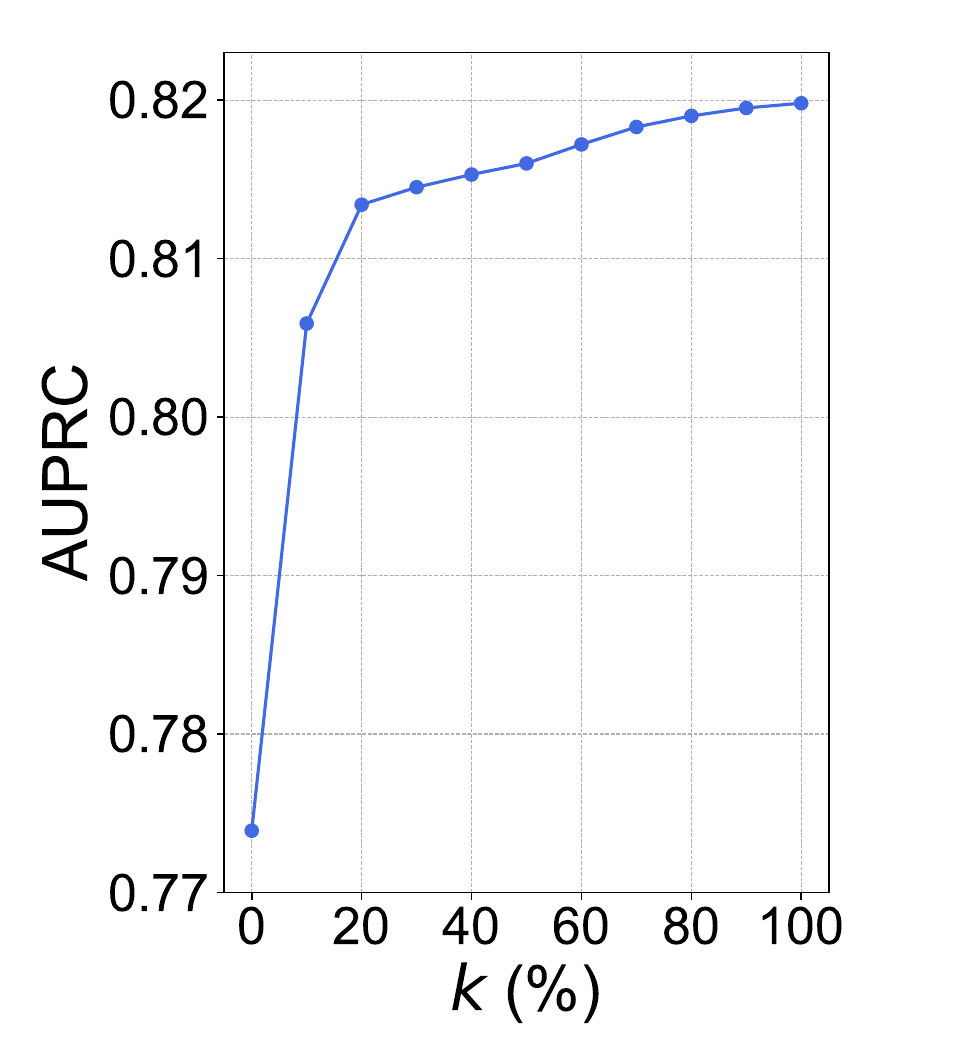}}
\subfigure[Impact of $m$.]{
\includegraphics[width=2.4cm]{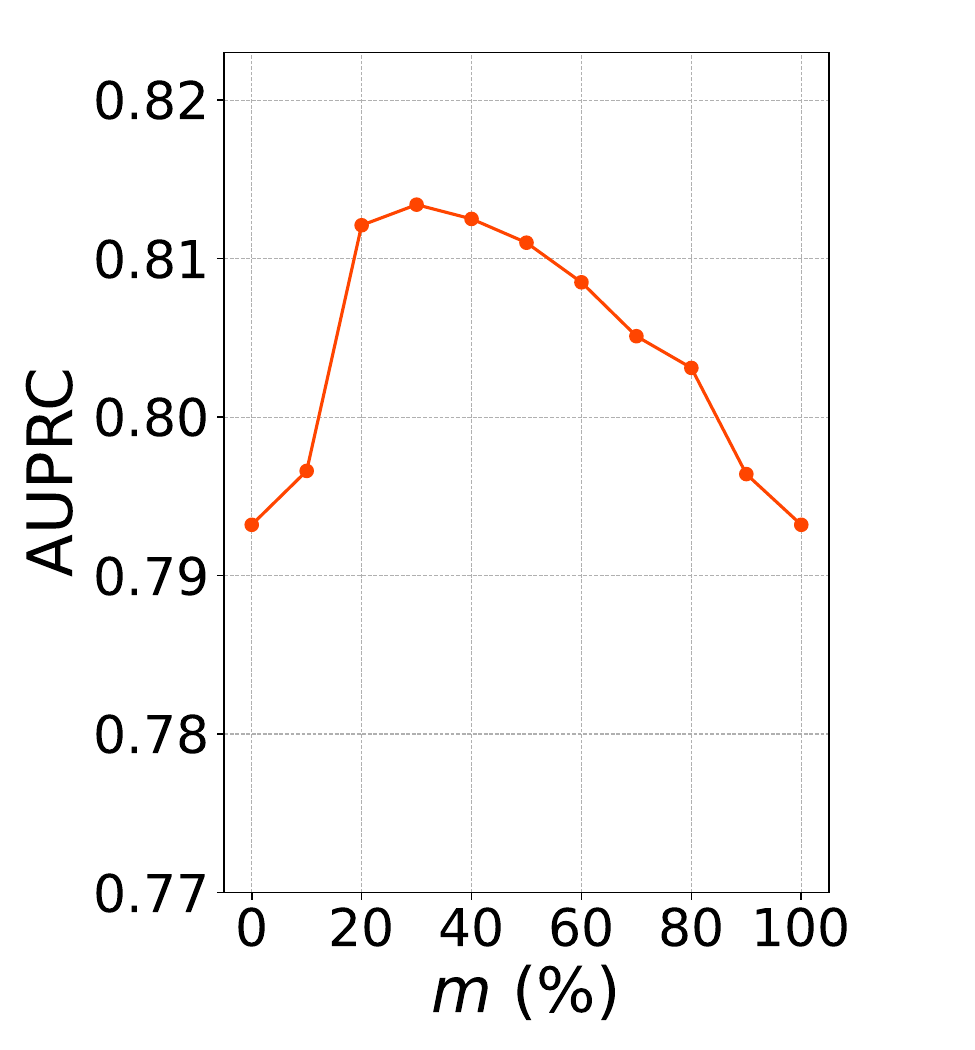}}
\subfigure[Impact of $\alpha$.]{
\includegraphics[width=2.4cm]{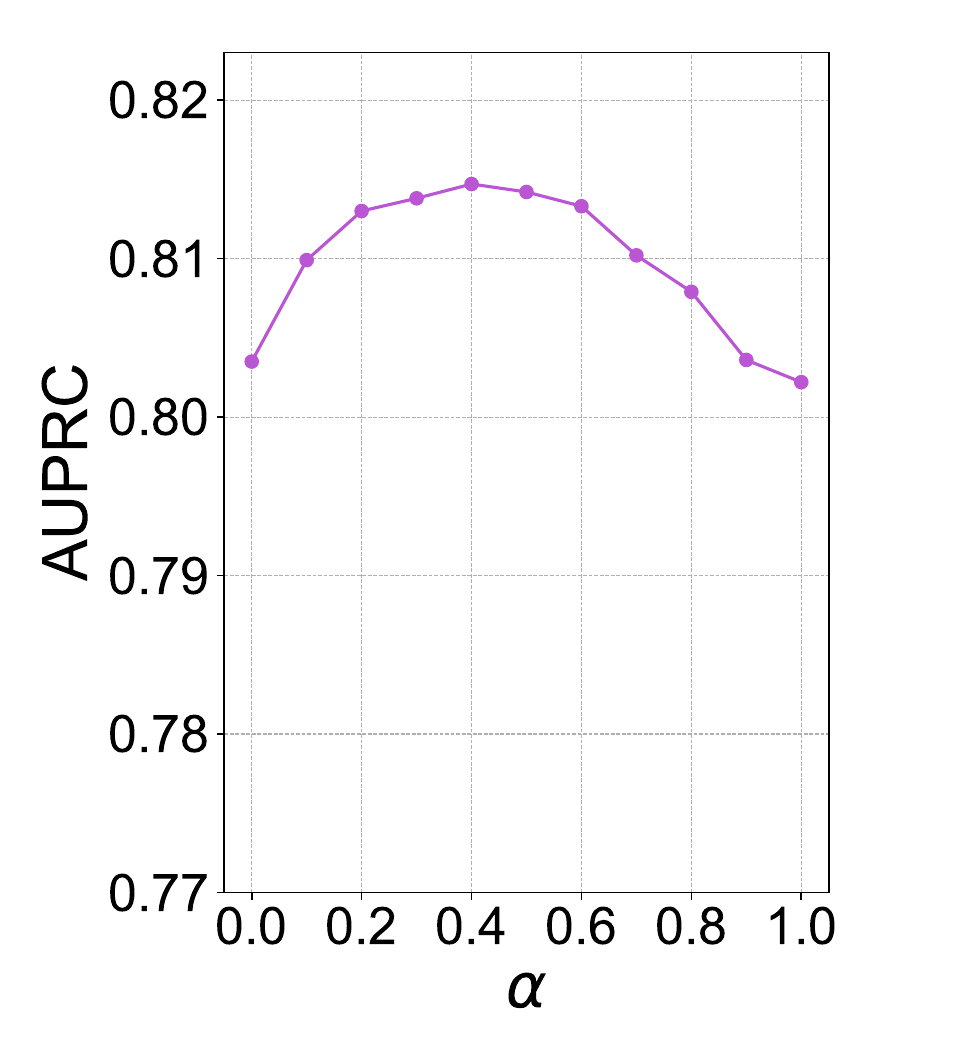}}
\vspace{-0.3cm}
\caption{Results of hyperparameter study.}
\vspace{-0.6cm}
\label{fig:hyper}
\end{figure}
In our method design, we introduce four hyperparameters: $k$ (proportion of text for LLM annotation), $m$ (proportion of emotionally intensive text), and $\alpha$ and $\beta$ (ratio between LLM analysis and original posts in mood course representation). The experimental results are shown in Figure~\ref{fig:hyper}.

\textbf{Impact of $k$.} As $k$ increases, the AUPRC monotonically increases. Annotating 20\% of the text with the LLM significantly improves AUPRC and is close to the performance when all texts are annotated. This demonstrates the rationality of our design for efficient implementation.

\textbf{Impact of $m$.} As $m$ increases, the AUPRC initially rises and then falls, suggesting a trade-off between capturing emotionally intensive posts and retaining sufficient information.

\textbf{Impact of $\alpha$ and $\beta$.} When $\alpha$:$\beta$ ratio is 2:3, the system achieves optimal performance.

\section{Practical Design Considerations}
\label{sec:practical_design}
Our research is motivated by the real-world need to address the global shortage of mental health services. Literature indicates an urgent need for effective, automated tools to assist with large-scale, low-cost early screening, complementing the efforts of human professionals~\cite{whitton2021mental,balcombe2021digital}. The design philosophy of DORIS is guided by deployability and considers several practical constraints:

\paragraph{Cost and Efficiency.} We designed an efficient filtering mechanism (Section~\ref{sec:efficient_annotation}) to significantly reduce expensive LLM API calls. As shown in the hyperparameter study (Appendix I), annotating only 20\% of the text achieves performance close to full annotation. This is a critical consideration for large-scale deployment.

\paragraph{Privacy and Flexibility.} In addition to using APIs, we validated our framework's effectiveness with locally deployed open-source LLMs (Appendix J). This directly addresses the core concerns of data privacy and security in real clinical settings and demonstrates deployment flexibility.

\paragraph{Explainability.} For medical applications, we view explainability not just as an academic pursuit but as a core prerequisite for gaining trust from clinicians and driving adoption. The hybrid design ensures that the system provides concrete evidence aligned with clinical practice.

\section{Prompt Instruction}
\label{sec:prompt}
The following prompt is used to instruct the LLM to identify depression symptoms within a given post:

\begin{center}
\begin{tcolorbox}[
    width=0.92\linewidth,
    colback=black!5!white,
    boxrule=0pt
    ]
    \textit{Assuming you are a psychiatrist specializing in depression. Given [text], please determine if this message includes any of the following states of the author:}
    
    \textit{A. Depressive mood B. Loss of interest/pleasure ... I. Thoughts of suicide.}
    
    \textit{If present, answer in the format of enclosed letters separated by commas, for example, (A, B, C). If none are present, respond with None.}
\end{tcolorbox}
\end{center}

The following prompt is used to instruct an LLM to synthesize a description of user $u$'s mood course, $T^{MC}$:
\begin{center}
\begin{tcolorbox}[
    width=0.92\linewidth,
    colback=black!5!white,
    boxrule=0pt
    ]
    \textit{"As a consulting psychiatrist, please conduct a longitudinal mood course analysis based on the following temporal sequence of personal expressions. For each entry, evaluate affect, emotional valence, and severity of mood states. Synthesize these observations into a clinical summary of mood progression, noting any patterns of persistence, fluctuation, or changes over time:}
    
    \textit{Time: $t_1$, Post: $p_1$, Time: $t_2$, Post: $p_2$, ..."}

\end{tcolorbox}
\end{center}

The following prompt is used to instruct the LLM to generate the explanation text, $T^{Exp}$:
 
\begin{center}
\begin{tcolorbox}[
    width=0.92\linewidth,
    colback=black!5!white,
    boxrule=0pt
    ]
    \textit{"Assuming you are a psychiatrist specializing in depression.}
    
    \textit{Here is a user's mood course: $T^{M C}$; below are posts from this user displaying symptoms of depression and the types of symptoms exhibited: ...; this user has been determined by an automated depression detection system to be depressed/normal.}
    
    \textit{Please consider the user's mood course and posts to generate an explanation for this judgment. Your explanation should be grounded in concrete evidence."}
\end{tcolorbox}
\end{center}

\section{Symptom Templates}
\label{sec:symptom_templates}
During the Diagnostic Criteria Feature Extraction phase, with the assistance of clinical psychologist and psychiatrists, we designed symptom templates to filter text with a high likelihood of depression through matching. Specifically, the templates we designed are as follows:

\paragraph{A. Depressed mood.} I feel low, unhappy, joyless, depressed, oppressed, gloomy, disappointed, melancholic, sad, distressed, heartbroken, a sense of loss, often feeling heavy-hearted, experiencing despair and despondency, always feeling sorrowful with an urge to cry, experiencing inner pain and emptiness.
\paragraph{B. Loss of interest/pleasure.} I have lost interest, feel indifferent, bored, unconcerned, lack enthusiasm, am unmotivated, have no interest in activities, am unmotivated, find almost everything uninteresting, lack motivation, find significantly reduced pleasure, cannot experience happiness, feel the world is dull, and cannot muster energy all day.
\paragraph{C. Weight loss or gain.} 
I experience reduced appetite, often feel full, lack of appetite, nausea, abnormal weight loss, difficulty swallowing, emaciation, loss of appetite, poor appetite, weight loss, or abnormal weight gain, sudden weight increase, unexplained weight gain.
\paragraph{D. Insomnia or hypersomnia.} 
I suffer from sleep disorders, depend on sleeping pills, often experience insomnia, have difficulty falling asleep, rely on sleep medication, frequently stay up late, struggle with sleep difficulties, and exhibit symptoms of insomnia, tossing and turning at night, or hypersomnia, oversleeping, sleep excess, prolonged sleep duration, or excessive sleepiness.
\paragraph{E. Psychomotor agitation or retardation.} I am neurotic, easily agitated, emotionally unstable, impatient, anxious, restless, mentally tense, irritable, often feeling mentally uneasy and agitated, fidgety, displaying impulsive and irritable behavior, and my emotions are easily out of control.
\paragraph{F. Fatigue} I feel fatigued, listless, exhausted, physically weakened, lacking in energy, dispirited, frequently tired, powerless, often feeling weary, unable to muster strength, feeling a heavy body, lacking in vitality and vigor, always feeling drowsy and lethargic.
\paragraph{G. Inappropriate guilt.} I have feelings of self-denial, lack of confidence, self-doubt, inferiority, disappointment, guilt, negative self-evaluation, self-blame, frequently belittle myself, feel incompetent and worthless, believe that I have achieved nothing and am a failure, feel disappointed in my expectations of myself and my family, often feel guilty and blame myself, thinking that everything is my fault.
\paragraph{H. Decreased concentration.} 
I experience slow thinking, difficulty concentrating, reduced judgment, memory decline, distractibility, indecision, scattered attention, difficulty thinking, lack of focus, difficulty paying attention, decreased cognitive ability, hesitancy in making decisions, often feeling mentally spaced out, unable to concentrate.
\paragraph{I. Thoughts of suicide.} 
I have a desire for death, self-harming behavior, suicidal thoughts, thoughts of ending my life, suicidal actions, thoughts of suicide, self-injury, recurring thoughts of death, suicidal tendencies, self-mutilation, cutting wrists with blades, jumping from heights to commit suicide, overdosing to commit suicide, making plans for suicide.

\section{Emotion Templates}
\label{sec:emotion_templates}
During the Mood Course Representation Construction phase, with the assistance of clinical psychologist and psychiatrists, we designed emotion templates. The purpose of emotion templates is to filter text with a high emotional intensity to generate the user's mood course. Specifically, the templates we designed are as follows:

\paragraph{1) Anger} I am angry, mad, agitated, annoyed, indignant, irritable, furious, disgusted, incensed, enraged, irritated, vexed, resentful, in a rage, glaring, shouting, screaming, insulting, hating, bellowing, outraged, ranting, detesting, fuming, and uncontrollably angry.
\paragraph{2) Disgust} I detest, loathe, disgust, abhor, hate, tire of, feel nauseated by, have a strong aversion to, despise, scorn, disdain, reject, find repugnant, utterly dislike, disdain, feel revulsion, despise, dislike intensely, abominate, have a strong displeasure, grow weary of, become impatient with, dismiss, look down upon, and utterly abhor.
\paragraph{3) Anxiety} I feel anxious, uneasy, worried, concerned, nervous, restless, panicked, fretful, afraid, uncertain, apprehensive, tense, jittery, indecisive, fearful, flustered, melancholic, frightened, apprehensive, full of doubts, brooding, terrified, distrustful, terrified, and on edge.
\paragraph{4) Happiness} I am happy, joyful, glad, blissful, merry, satisfied, delighted, elated, pleased, laughing, cheerful, excited, jubilant, optimistic, enthusiastic, cheerful, uplifted, exuberant, overjoyed, jubilant, with a smile on my face, pleasantly surprised, beaming with joy, and my heart blooms with happiness.
\paragraph{5) Sadness} I am sad, sorrowful, melancholic, in pain, lost, pessimistic, tearful, grieving, mournful, depressed, suicidal, heartbroken, devastated, upset, crying, deeply saddened, disconsolate, dejected, lamenting, desolate, gloomy, weeping bitterly, desperate, heartbroken, indignant.

\end{document}